\documentclass[10pt,twocolumn,letterpaper]{article}

\usepackage{cvpr}   

\usepackage[table]{xcolor}

\usepackage{soul}
\setuldepth{foobar}

\usepackage{graphicx}
\usepackage{booktabs}
\usepackage{multirow}
\usepackage{algorithm}
\usepackage{algpseudocode}
\usepackage{amssymb} 

\usepackage{tikz}
\usetikzlibrary{arrows.meta, positioning, shapes.geometric}

\newcommand{\valuestd}[2]{#1{\scriptsize $\pm$#2}}

\usepackage{tcolorbox}
\tcbuselibrary{breakable,skins}

\newcommand\blfootnote[1]{%
  \begingroup
  \renewcommand\thefootnote{}\footnote{#1}%
  \addtocounter{footnote}{-1}%
  \endgroup
}

\definecolor{cvprblue}{rgb}{0.21,0.49,0.74}
\usepackage[pagebackref,breaklinks,colorlinks,citecolor=cvprblue]{hyperref}

\def\paperID{} 
\def\confName{CVPR}
\def\confYear{2026}

\title{ArtHOI: Taming Foundation Models for Monocular 4D Reconstruction of \\ Hand-Articulated-Object Interactions }

\author{Zikai Wang$^{1}$ \quad Zhilu Zhang$^{1, *}$ \quad Yiqing Wang$^{2}$ \quad Hui Li$^{1}$ \quad Wangmeng Zuo$^{1}$ \\
{
 $^1$Harbin Institute of Technology \quad
 $^2$Shanghai Jiao Tong University
}
}

\begin{document}

\twocolumn[{%
\renewcommand\twocolumn[1][]{#1}%
\maketitle

\begin{center}
  \includegraphics[width=0.96\textwidth]{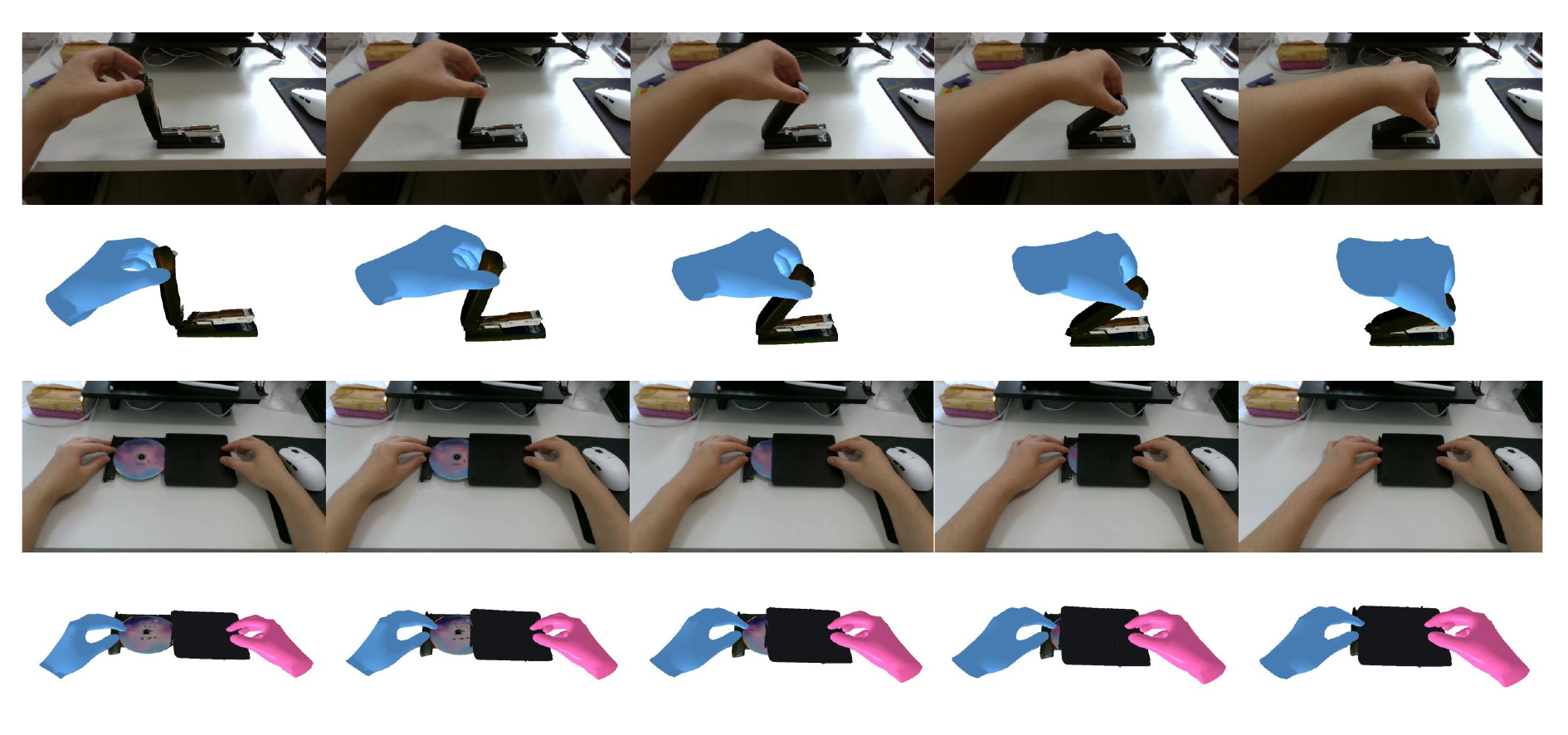}
  \vspace{-6mm}
  \captionof{figure}{%
    Given a monocular RGB video sequence of hands interacting with an unknown articulated object, our method, ArtHOI, reconstructs 4D human-object interactions (HOI) without any pre-defined object templates or multi-view scan initialization. 
    Here we show two examples of input videos and the reconstructed HOI results.
  }
  \label{fig:teaser}
\end{center}
}]

\blfootnote{$^*$ Corresponding author. Email: cszlzhang@outlook.com}

\begin{abstract}
  Existing hand-object interactions (HOI) methods are largely limited to rigid objects, while 4D reconstruction methods of articulated objects generally require pre-scanning the object or even multi-view videos. It remains an unexplored but significant challenge to reconstruct 4D human-articulated-object interactions from a single monocular RGB video. Fortunately, recent advancements in foundation models present a new opportunity to address this highly ill-posed problem. To this end, we introduce ArtHOI, an optimization-based framework that integrates and refines priors from multiple foundation models. Our key contribution is a suite of novel methodologies designed to resolve the inherent inaccuracies and physical unreality of these priors. In particular, we introduce an Adaptive Sampling Refinement (ASR) method to optimize object's metric scale and pose for grounding its normalized mesh in world space. Furthermore, we propose a Multimodal Large Language Model (MLLM) guided hand-object alignment method, utilizing contact reasoning information as constraints of hand-object mesh composition optimization. To facilitate a comprehensive evaluation, we also contribute two new datasets, ArtHOI-RGBD and ArtHOI-Wild. Extensive experiments validate the robustness and effectiveness of our ArtHOI across diverse objects and interactions. Project: 
  \href{https://arthoi-reconstruction.github.io}{https://arthoi-reconstruction.github.io}.

\end{abstract}    
\section{Introduction}
\label{sec:intro}

Hand-Object Interactions (HOI) reconstruction~\cite{gkioxari2018detecting, hasson19_obman, chen2021joint, chen2022alignsdf, fan2024hold, liu2025easyhoi, wang2025magichoi, shan2020understanding,yang2025physworld} aims at obtaining a physically plausible 3D representation of hands, objects, and their interplay from visual observations. It plays a crucial role in various applications, including human behavior analysis~\cite{kikuchi2024self}, robotic manipulation~\cite{zhou2025you, kerr2024rsrd, beingbeyond2026beingh05}, and augmented reality~\cite{salvato2022predicting}.

Early works usually require predefined object templates~\cite{gkioxari2018detecting, brahmbhatt2020contactpose, cao2021reconstructing, chen2021joint, grady2021contactopt, huang2022reconstructing} or category-specific knowledge~\cite{chao2021dexycb, yang2022oakink, li2020detailed}, which limited their applicability to unconstrained, wild scenarios. 
While recent template-free and category-independent methods~\cite{fan2024hold, liu2025easyhoi, wang2025magichoi, swamy2025host3r, aytekin2025follow} have demonstrated improved generalization, they largely operate under the assumption of rigid objects.
Furthermore, we also note that significant progress~\cite{kerr2024rsrd, liu2025videoartgs, le2024articulate, yu2025artgs, peng2025itaco, yu2025part, song2024reacto, wu2025predict, liu2023paris, liu2025surveymodelinghumanmadearticulated} has been made in 4D articulated object reconstruction through optimization-based~\cite{liu2023paris, kerr2024rsrd, liu2025videoartgs, yu2025artgs, peng2025itaco} and learning-based~\cite{jiang2022ditto, mokhtar2024centerart} techniques, but these methods typically rely on pre-scanning objects (for canonical shape)~\cite{kerr2024rsrd, wu2025predict, peng2025itaco} or even multi-view videos~\cite{liu2025building, yu2025part}.
Consequently, in uncontrolled environments where articulated objects (\eg, scissors, eyeglasses, and laptops) are manipulated naturally, HOI reconstruction from monocular videos remains an unexplored challenge.

It is an inherently ill-posed task due to limited visual cues and frequent occlusions, making the design of an effective and robust method non-trivial. 
In contrast, humans can effortlessly perceive such complex interactions, a capability that stems from accumulated knowledge and experience.
Drawing inspiration from this human faculty, we argue that a promising solution lies in leveraging the rich priors of various foundation models.
Specifically, these models can provide critical geometric, motion, and semantic information. 
For instance, image-to-3D~\cite{hong2023lrm, li2023instant3d, xu2024instantmesh, long2023wonder3d, lai2025hunyuan3d} can recover 3D shape of an articulated object, and pose estimation~\cite{wen2024foundationpose,nguyen2024gigaPose} can compute its 6D transformation relative to the camera. 
Furthermore, depth estimation~\cite{piccinelli2024unidepth, chen2025video} and tracking~\cite{karaev2025cotracker3, xiao2025spatialtracker} can offer metric geometry and motion cues, respectively.
For the hand, specialized models~\cite{pavlakos2024reconstructing,potamias2025wilor} can reconstruct its 3D mesh. Multimodal Large Language Models (MLLMs)~\cite{wang2024qwen2, comanici2025gemini} can infer the interaction state between the hand and the object.

Nevertheless, a naive integration of these foundation models is prone to failure, as their individual predictions sometimes contain inaccuracies and some are not inherently grounded in the physical reality.
In particular, image-to-3D models typically generate geometry in a normalized, object-centric coordinate system, lacking the metric scale required to determine the object's true pose in world space. Furthermore, even if the 4D representation of the object is accurately reconstructed, simply composing it with a hand mesh often leads to physically implausible results, such as interpenetration or disjointed contact, due to spatial misalignments between the two.

To address these issues, we propose ArtHOI, a novel framework for reconstructing 4D hand-articulated-object interactions from a monocular video, which optimizes the inconsistency and mismatch problems while collaboratively leveraging priors of foundation models.
In particular, firstly, we propose an Adaptive Sampling Refinement (ASR) method to estimate the metric scale and 6-DoF pose of the canonical articulated object. It is used to recover 3D mesh in world space from the generated normalized one and prepare the object motion reconstruction.
Secondly, for hand-object mesh composition, we elaborate the prompts for MLLM to infer frame-wise contact states and fingers. The contact information is then used as optimization constraints to jointly refine the object scale and hand pose, improving their spatial alignment.

Specifically, the ArtHOI pipeline mainly comprises four stages: data preprocessing, canonical object mesh reconstruction, part-wise object motion reconstruction, and hand-object alignment.
First, the preprocessing stage leverages foundation vision models to extract hand and object masks, metric depths, camera parameters, \etc. A video inpainting model is applied to restore the object regions occluded by the hand.
Second, we deploy an image-to-3D model to generate a normalized 3D mesh from the inpainted object. This mesh is then scaled and oriented in world space using our proposed ASR method.
Third, we initialize coarse motion trajectories for each object part using a dense tracking model. These trajectories, along with part visibilities, are then used to solve for the per-part  $\mathrm{SE}(3)$ transformations over time.
Finally, hand reconstruction is performed, and hand-object interaction is refined via our MLLM-guided alignment method.

To facilitate a more comprehensive evaluation, we supplement the existing RSRD~\cite{kerr2024rsrd} dataset with two new benchmarks: ArtHOI-RGBD, comprising RGBD videos captured with a RealSense camera, and ArtHOI-Wild, consisting of challenging videos collected from the internet. Experiments demonstrate our ArtHOI effectively reconstructs physically plausible 4D HOI across diverse objects and interactions.
Notably, our method achieves superior performance even when compared to RSRD~\cite{kerr2024rsrd} that relies on pre-scanned object geometry as input.

Our contributions are summarized as follows:
\begin{itemize}
    \item We introduce ArtHOI, an optimization-based framework that reconstructs 4D hand-articulated-object interactions from monocular videos via integrating and refining priors from multiple foundation models.
    
    \item We propose an Adaptive Sampling Refinement (ASR) method to optimize object's metric scale and pose, which serves object mesh reconstruction in the world space.
    
    \item We propose an MLLM-guided hand-object alignment method that performs contact reasoning for constrainting hand-object mesh composition.

    \item We conduct extensive experiments on existing and newly introduced challenging datasets, which demonstrated the superior robustness and effectiveness of our method across diverse objects and interactions.

\end{itemize}

\section{Related Works}

\subsection{Hand-Object Interaction Reconstruction}
Reconstructing hand-object interaction (HOI) from monocular RGB images or video~\cite{shan2020understanding, cao2021reconstructing, chao2021dexycb, chen2021joint, huang2022reconstructing, chen2022alignsdf, chen2023gsdf, huo2024wildhoi, fan2024hold, swamy2025host3r, aytekin2025follow, on2025bigs, liu2025easyhoi, wen2025reconstructing, wang2025magichoi, yang2025physworld} is intrinsically difficult due to severe occlusions and depth ambiguities~\cite{fan2024hold, wang2025magichoi, aytekin2025follow}. Early solutions addressed this by assuming known object templates~\cite{gkioxari2018detecting, brahmbhatt2020contactpose, cao2021reconstructing, chen2021joint, grady2021contactopt, huang2022reconstructing} or pretraining on small-scale 3D object datasets~\cite{yang2022oakink, chao2021dexycb, shan2020understanding}. More recent, model-free approaches exploit priors from large reconstruction or foundation models: some employ pretrained large reconstruction models (LRMs) to obtain an initial object shape~\cite{liu2025easyhoi, wen2025reconstructing}, while others use novel-view synthesis~\cite{wang2025magichoi} to recover geometry under sparse view inputs. Nonetheless, many of these methods are restricted to image inputs, rigid-object assumptions, or static contact states~\cite{liu2025easyhoi, wen2025reconstructing, aytekin2025follow, huang2022reconstructing} during optimization; consequently they do not handle dynamic interactions or complex articulated objects well. Importantly, rich real-world priors can serve not only for shape initialization but also for articulated motion analysis and dynamic contact reasoning. By fully exploiting such priors from multiple foundation models~\cite{chen2025video, ravi2024sam, lai2025hunyuan3d, karaev2025cotracker3, wen2024foundationpose, Qwen-VL}, our work advances 4D reconstruction of dynamic hand-articulated-object interactions from casual monocular videos.

\subsection{4D Reconstruction of Articulated Object}
Reconstructing real-world articulated objects from limited input remains a challenging problem. Earlier methods typically require 3D point-cloud inputs~\cite{liu2023building, jiang2022ditto, nie2022structure} or multi-view observations~\cite{huang2012occlusion, zhang2021strobenet, yu2025part}; constrained by these requirements, they usually rely on synthesized~\cite{mo2019partnet, liu2022akb, geng2023gapartnet} or laboratory-captured datasets~\cite{kerr2024rsrd, fan2023arctic} and thus do not generalize well to in-the-wild data. Recent work has begun to reconstruct articulated objects from monocular RGB video captured in the wild~\cite{song2024reacto, kerr2024rsrd, peng2025itaco, wu2025predict, le2024articulate, werby2025articulated, liu2025videoartgs}, achieving promising results by combining flexible 3D representations with rich priors from foundation models such as DINOv2~\cite{oquab2023dinov2}, SAM~\cite{ravi2024sam}, and dense tracking models~\cite{karaev2025cotracker3, xiao2025spatialtracker, tapip3d}. 
However, most of these approaches assume an initial pre-scanned sequence (object observed from surrounding viewpoints)~\cite{kerr2024rsrd, peng2025itaco, liu2025videoartgs} 
or depend on predefined part libraries~\cite{le2024articulate, xu2022unsupervised}. 
This initialization provides full-view coverage and a static geometry prior but is impractical for casual capture. Moreover, existing methods typically model only the articulated object and ignore the interacting hand present in real manipulation videos. While effective in controlled settings, these limitations hinder applicability to natural interaction scenarios. By leveraging and coordinating multiple foundation-model priors, our approach relaxes these restrictions, enables joint reconstruction of hands and articulated objects from casually captured monocular interaction videos.

\section{Method}
\label{method}

\label{sec:pipeline}

\begin{figure*}[ht!]
  \centering
  \includegraphics[width=0.99\linewidth]{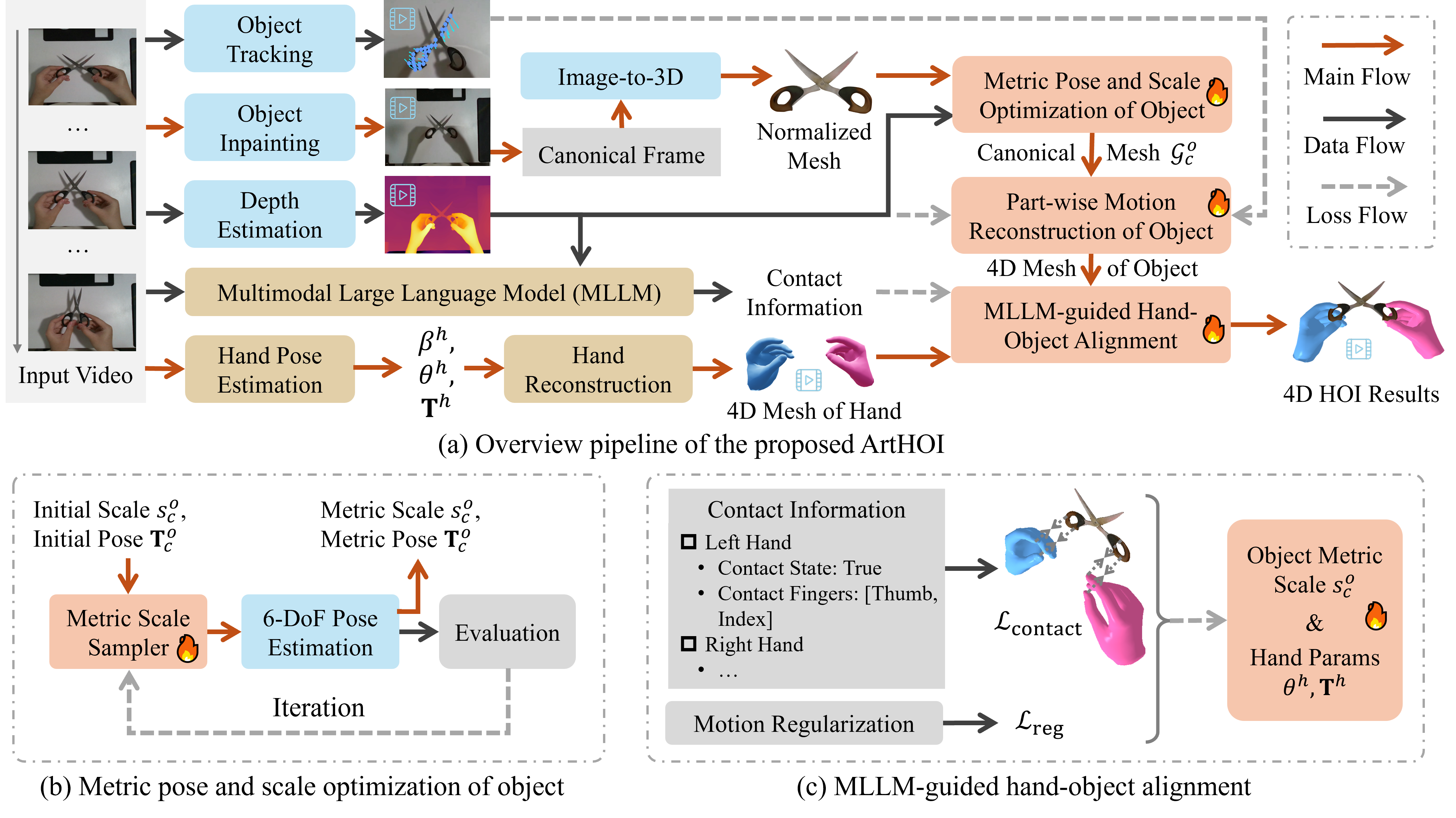}
  \vspace{-2mm}
  \caption{Pipeline of our ArtHOI. ArtHOI is an optimization-based framework (see subfigure (a)) that integrates and refines priors from multiple foundation models for monocular 4D reconstruction of human-articulated-object interactions. In particular, the proposed object's metric scale and pose optimization (see subfigure (b)) recovers 3D mesh in world space from a normalized one, while MLLM-guided hand-object alignment method (see subfigure (c)) promotes physically plausible hand-object mesh composition.}
  \label{fig:framework}
  \vspace{-2mm}
\end{figure*}

Our ArtHOI framework mainly consists of four stages.
In Sec.~\ref{sec:preproc}, we employ a set of foundation models to preprocess the input video and extract multi-dimensional priors. 
Sec.~\ref{sec:cano_init} constructs a canonical representation of the articulated object, including its mesh, metric scale, and 6-DoF global pose.
In Sec.~\ref{sec:part_motion}, we estimate part-wise $\mathrm{SE}(3)$ motion trajectories from dense tracking priors via an occlusion-aware optimization. 
Finally, Sec.~\ref{sec:hoalign} integrates a hand reconstruction model to recover 4D hand mesh, and employs MLLM-guided HOI alignment optimization that resolves spatial mismatches between the reconstructed hands and the object. The pipeline of ArtHOI can be seen in Fig.~\ref{fig:framework}.

\subsection{Data Preprocessing}
\label{sec:preproc}

Given a monocular video $\mathcal{V}=\{\mathbf{I}_i\}_{i=1}^N$ of $N$ RGB frames, we first apply several foundation vision models to extract informative priors. Object masks $\{\mathbf{M}_i\}_{i=1}^N$ and human masks are obtained using a video segmentation model~\cite{ravi2024sam}. Metric depth maps $\{\mathbf{D}_i\}_{i=1}^N$ and camera intrinsics $\mathbf{K}$ of the input video are estimated with a monocular depth estimator~\cite{chen2025video}. To mitigate hand-object occlusions, we apply a video inpainting model~\cite{li2025diffueraser} to remove the human from the input video, producing an inpainted video $\mathcal{V'}=\{\mathbf{I}'_i\}_{i=1}^N$ containing only the object. The inpainted video is further processed with the same preprocessing pipeline to extract object-only masks $\{\mathbf{M}'_i\}_{i=1}^N$ and depth maps $\{\mathbf{D}'_i\}_{i=1}^N$.

We then leverage priors from a large image-to-3D reconstruction model, HunYuan3D~\cite{lai2025hunyuan3d}, to recover the complete geometry of the articulated object. 
Specifically, let the inpainted canonical frame be denoted by $\mathbf{I}'_c$, we extract the object image from $\mathbf{I}'_c$ using its mask $\mathbf{M}^{\prime o}_c$, and feed the cropped object image into HunYuan3D to obtain its 3D mesh.

\begin{algorithm}[ht!]
\caption{Adaptive Sampling Refinement (ASR)}
\label{algo:asr}
\begin{algorithmic}[1]
\Require Normalized object mesh $\mathcal{G}^o$; RGB $\mathbf{I}'_c$, depth $\mathbf{D}'_c$ and mask $\mathbf{M}'^{\,o}_c$ of canonical frame; camera intrinsics $\mathbf{K}$; number of iterations $J$; initial sampling range $\delta$
\Ensure Metric scale $s_c^o$ and pose $\mathbf{T}_c^o$ of canonical object, scaled canonical object mesh $\mathcal{G}_c^o$

\State $s^o_{\textrm{coarse}} \gets \textsc{CoarseScaleEstimation}(\mathcal{G}^o, \mathbf{D}'_c, \mathbf{M}'^{\,o}_c)$
\State $(\mathcal{L}_{\mathrm{best}}, j_{\mathrm{best}}) \gets (-\infty, 0)$
\For{$j = 1$ \textbf{to} $J$}
    \If{$j_{\mathrm{best}} < \frac{j}{2}$} 
        \State $\delta \gets 2 \delta$ \Comment{Adaptively expand the range}
    \EndIf
    \State $\hat{s_c^o} \gets s^o_{\textrm{coarse}} \cdot \textsc{RandomSample}(-\delta,\delta)$
    \State $\hat{\mathcal{G}}^o_c \gets \textsc{Scale}(\mathcal{G}^o, \hat{s_c^o})$
    \State $\hat{\mathbf{T}}_c^o \gets \textsc{FoundationPose}(\hat{\mathcal{G}}^o, \mathbf{I}'_c, \mathbf{D}'_c, \mathbf{M}'^{\,o}_c, \mathbf{K})$
    \State $\hat{\mathbf{M}}^{o}_c \gets \textsc{RenderSilhouette}(\hat{\mathbf{T}}_c^o \cdot \hat{\mathcal{G}}^o, \mathbf{K})$
    \State $\mathcal{L}_{\mathrm{iou}} \gets \textsc{iou}(\hat{\mathbf{M}}^{o}_c, \mathbf{M}'^{\,o}_c)$
    \If{$\mathcal{L}_{\mathrm{iou}} > \mathcal{L}_{\mathrm{best}}$}
        \State $s_c^o \gets \hat{s_c^o}$, $\mathbf{T}_c^o \gets \hat{\mathbf{T}}_c^o$, $\mathcal{L}_{\mathrm{best}} \gets \mathcal{L}_{\mathrm{iou}}$, $j_{\mathrm{best}} \gets j$
    \EndIf
    \EndFor
\State $\mathcal{G}^o_c \gets \textsc{Scale}(\mathcal{G}^o, s_c^o)$
\State \Return $s_c^o, \mathbf{T}_c^o$, $\mathcal{G}^o_c$
\end{algorithmic}
\end{algorithm}

\subsection{Metric Pose and Scale Optimization of Object}
\label{sec:cano_init}

Here we align the normalized mesh produced by HunYuan3D with other priors (including the estimated metric depth $\mathbf{D}^{\prime o}_c$ and object mask $\mathbf{M}^{\prime o}_c$) to obtain a metric canonical mesh in world space. It is achieved by optimizing metric scale $s_c^o$ and 6-DoF pose $\mathbf{T}_c^o$ of the object.

A natural option is to directly apply a state-of-the-art 6-DoF pose estimator, \eg, FoundationPose~\cite{wen2024foundationpose}, on the inpainted frame $\mathbf{I}'_c$ with $\mathbf{D}^{\prime o}_c$ and $\mathbf{M}^{\prime o}_c$. However, while FoundationPose performs well when given accurate metric depth and a metric-scaled ground-truth mesh, its performance degrades notably in our setting due to the inconsistencies between the generated mesh and inaccurate depth, leading to poor or unstable predictions.

To reconcile these heterogeneous priors, we introduce  an \textbf{Adaptive Sampling Refinement (ASR)} method. ASR first computes a coarse scale estimate for the normalized mesh by using back-projected metric depth, then iteratively samples candidate scales from an adaptive range around initial estimate. For each sampled candidate scale, ASR queries FoundationPose to produce pose hypothesis, and evaluates each hypothesis by rendering the posed mesh and matching the rendered silhouette against the preprocessed object mask. The sampling range is adaptively adjusted based on recent refinement progress: if no improvement is observed in recent iterations, the sampling range is expanded; otherwise it is kept unchanged. The algorithm selects the final scale and pose with the best rendered feedback. By searching metric scales and validating pose hypotheses, ASR robustly coordinates the normalized mesh, noisy depth, and pose predictions to yield a reliable metric scale  $s_c^o$  and pose $\mathbf{T}_c^o$. The detailed procedure is given in Algorithm~\ref{algo:asr}.

\subsection{Part-wise Motion Reconstruction}
\label{sec:part_motion}

To effectively exploit both spatial and temporal cues while handling part-wise occlusions, we leverage dense tracking priors~\cite{karaev2025cotracker3, xiao2025spatialtracker} to obtain coarse part motions and then optimize per-part SE(3) transformations over time.

Concretely, denote part masks of $i$-th frame  as $\{\mathbf{M}'^{\, p_k}_i\}_{k=1}^K$, we first partition the canonical object mesh $\mathcal{G}^o_c$ into parts by applying PartField~\cite{partfield2025} to group vertices and using these masks for partition.
We run CoTracker~\cite{karaev2025cotracker3} on the inpainted video $\mathcal{V'}$ to produce temporally coherent point tracks together and per-point visibilities. For the $k$-th part, we sample $Q$ query pixels inside its mask $\mathbf{M}'^{\, p_k}$ and track sampled queries using CoTrackerV3~\cite{karaev2025cotracker3}, which outputs a 2D point trajectory together with a per-frame visibility indicator. Then we lift them to 3D using the depth map $\mathbf{D}'_i$, yielding the 3D track and visibility pair  $(\mathbf{z}_{i,q}^{k},\, v_{i,q}^{k})$, where $v_{i,q}^{k}\in\{0,1\}$. Therein, outlier tracks are removed by a lightweight post-processing operation.

We then optimize per-part $\mathrm{SE}(3)$ transformations across frames, denoted $\{\mathbf{T}_{i}^{p_k}\}_{i=1}^N$, by enforcing consistency with 3D tracking priors under visibility constraints. For the $k$-th part in $i$-th frame, let $\,\mathbb{S}\,$ be a set of sampled reference frames, the tracking loss is 
\begin{equation}
\label{eq:track_loss}
\mathcal{L}_{\mathrm{track}}\ \;=\;
\sum_{j\in \mathbb{S}}\; \sum_{q \in \mathbb{W}_{i,j}^k}
\big\| \mathbf{z}_{j,q}^{k} \;-\; (\mathbf{T}_i^{p_k})^{-1}\mathbf{T}_j^{p_k}\,\mathbf{z}_{i,q}^{k} \big\|,
\end{equation}
where $\mathbb{W}_{i,j}^k=\{q \mid v_{i,q}^k=1 \land v_{j,q}^k=1\}$ is the set of tracks visible in both frames $i$ and $j$.
To regularize the temporal motion dynamics, we further apply a smoothness constraint:
\begin{equation}
\mathcal{L}_{\mathrm{smooth}} =
\sum_{i=2}^{N-1} 
\big\| \Delta^2 \mathbf{T}_i^{p_k} \big\|.
\end{equation}
where $\Delta^2$ denotes the discrete second-order difference operator applied along the temporal dimension, \ie, $\Delta^2 \mathbf{T}_i^{p_k} = \mathbf{T}_{i+1}^{p_k} - 2\mathbf{T}_i^{p_k} + \mathbf{T}_{i-1}^{p_k}$.

Finally, the overall objective for part-wise motion optimization is formulated as
\begin{equation}
\vspace{-2mm}
\mathcal{L_{\mathrm{motion}}} = 
\mathcal{L}_{\mathrm{track}} + 
\lambda_{\mathrm{smooth}} \mathcal{L}_{\mathrm{smooth}}.
\end{equation}

\begin{figure*}[t]
  \centering
  \includegraphics[width=0.99\linewidth]{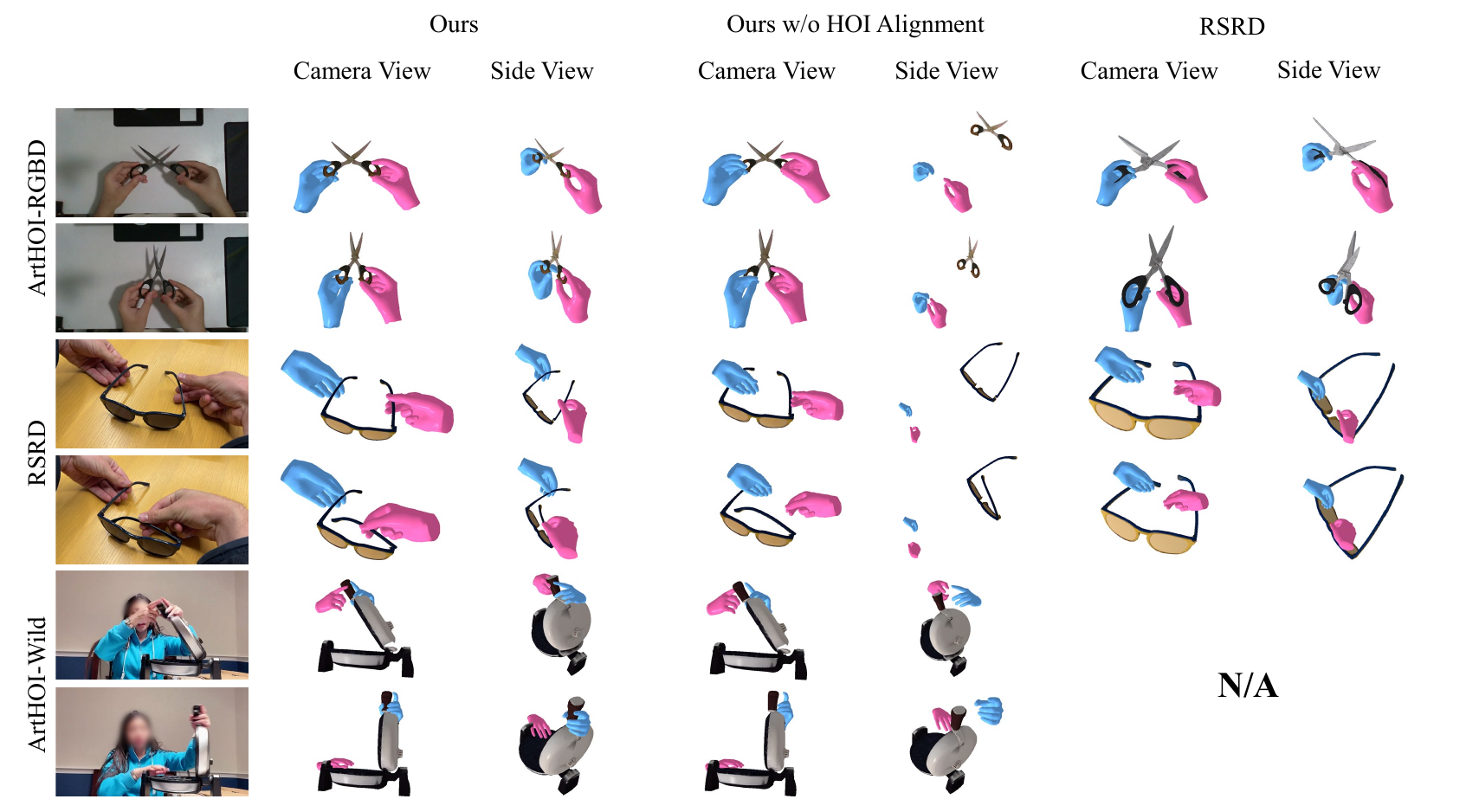}
  \vspace{-1mm}
  \caption{This gallery showcases the results of our hand-articulated-object reconstruction on three data sources: ArtHOI-RGBD, RSRD and ArtHOI-Wild.(more results in the supp.). The first column shows sampled input frames. We present the camera view and a side view to display the reconstructed HOI meshes. Hand reconstructions for RSRD are produced using the same WiLoR model as ours for a fair comparison. Note that RSRD is unable to process the video from ArtHOI-Wild, as it requires an object surrounding scan that is unavailable for internet videos. }
  \label{fig:fig_hoalign}
\vspace{-3mm}
\end{figure*}

\subsection{MLLM-guided Articulated HOI Alignment}
\label{sec:hoalign}

We employ the off-the-shelf hand pose estimator WiLoR~\cite{potamias2025wilor} to reconstruct MANO-based 4D hands, parameterized by articulated hand joint poses $\{\theta_i^h\}_{i=1}^N \in \mathbb{R}^{N \times 45}$, hand shape $\beta^h \in \mathbb{R}^{10}$ and global transformation $\{\mathbf{T}^h_i\}_{i=1}^N$.
To handle missing or unreliable predictions due to occlusions, we apply spherical linear interpolation (SLERP) on hand pose and global transformation to temporally smooth and fill in the hand poses and transformations.

Separated reconstruction of 4D articulated objects and hands often produces spatio-temporal misalignments due to inconsistencies among different priors, motivating a joint optimization for articulated HOI. To enable dynamic interaction reasoning, we leverage Multimodal Large Language Models (MLLMs)~\cite{wang2024qwen2, comanici2025gemini} to infer contact information, including the binary contact state and contacting fingers for each frame, leveraging their rich real-world priors and multimodal reasoning capabilities. However, naively querying MLLMs for contact estimation is insufficient: diverse camera viewpoints often lead to left–right hand confusion, while limited RGB cues make it difficult to distinguish true physical contact from mere proximity.

To mitigate these issues, we design a structured prompting strategy. First, we ask the MLLM to determine the camera perspective (egocentric vs. exocentric) of the video and incorporate this information into subsequent contact queries. Next, we infer frame-wise contact information—including hand laterality, binary contact state, and contacting fingers—by iteratively querying each frame with the constructed prompt. To provide richer contextual cues, we concatenate $k$ neighboring RGB frames along with their colorized depth maps to form a large image prompt. This pipeline yields more reliable frame-wise estimates for subsequent optimization. We denote the set of frames where the hand is in contact with the object as $\mathbb{C}$, and the set of contacting fingers in the $i$-th frame as $\mathbb{Y}_i$.

We leverage the retrieved contact information as frame-wise constraints to guide 4D hand-object interaction alignment. Our optimization follows a two-stage procedure. Given that WiLoR~\cite{potamias2025wilor} provides reliable metric scale priors of hand, while estimated depth may remain ambiguous, the first stage optimizes only object scale $s^o_c$ to align with the hand. In the second stage, we fix the optimized object scale and jointly refine the hand pose parameters $\theta_i^h$ and global transformations $\mathbf{T}_i^h$ to further enhance the spatial consistency between the interacting hand and object.

Let $\mathbb{T}_i$ denote the set of MANO fingertip vertices corresponding to $\mathbb{Y}_i$. The contact loss $\mathcal{L}_{\mathrm{contact}}$  minimizes the distance from each fingertip to the closest point from object mesh $\mathcal{G}^o_i$. It can be written as
\begin{equation}
\label{eq:contact}
\mathcal{L}_{\mathrm{contact}}
= \sum_{i \in \mathbb{C}} \;\sum_{\mathbf{v}_t \in \mathbb{T}_i}
\; \min_{\mathbf{v}_o \in \mathcal{G}^o_i} \; \big\| \mathbf{v}_o - \mathbf{v}_t \big\|_2 .
\end{equation}

To further regularize the optimization, we introduce a motion regularization term $\mathcal{L}_{\mathrm{reg}}$ over hand parameters $\theta_i^h$ and global transforms $\mathbf{T}_i^h$. This term combines an acceleration prior on $\mathbf{T}_i^h$ and an $\ell_1$ penalty between the optimized pose with the
initial pose $\theta_i^{h,init}$, \ie,
\begin{equation}
\label{eq:reg}
\mathcal{L}_{\mathrm{reg}} =
\lambda_{\mathrm{acc}} \;\big\| \Delta^2 \mathbf{T}^h \big\|_2
+
\lambda_{\theta} \;\sum_{i=1}^N \big\| \theta_i^h - \theta_i^{h,init} \big\|_1.
\end{equation}
Finally, the overall HOI alignment loss can be written as
\begin{equation}
\label{eq:hoi_loss}
\mathcal{L}_{\mathrm{hoi}}
=  \mathcal{L}_{\mathrm{contact}}
+ \mathcal{L}_{\mathrm{reg}}.
\end{equation}

\section{Experiments}
\label{sec:experiments}

\subsection{Datasets}

We capture five demonstration sequences of common articulated objects using an Intel RealSense stereo camera at $1280\times720$ and 30 FPS with accurate metric depth; we denote this collection as \textbf{ArtHOI-RGBD}. In addition, we collect eight in-the-wild clips from internet sources and smartphone recordings, denoted \textbf{ArtHOI-Wild}. Experiments are performed on these two collections, and we additionally evaluate on nine videos from the RSRD dataset, as well as a three-object subset of ARCTIC~\cite{fan2023arctic}, covering diverse objects and manipulation scenarios.

Because the ground-truth depth in ArtHOI-RGBD provides only partial surface observations, we develop a 3D annotation tool (built on Viser~\cite{yi2025viser}) to label part-wise object motions across frames for all five videos and four RSRD videos under the help of depth maps as geometric guidance. To obtain complete object geometry, we additionally capture a surrounding scan for each object to reconstruct full ground-truth meshes (used by RSRD). We also annotate hand-object contact states for all used videos.

\subsection{Implementation Details}
Our system can be implemented on an NVIDIA A6000 GPU, with a total computation time of $\sim$1 hour for a monocular video input with 100 frames under $960\times540$ resolution. We use Video-Depth-Anything~\cite{chen2025video} for depth estimation with UnidepthV2~\cite{piccinelli2024unidepth} for metric scaling and camera parameter recovery. We adopt Segment-Anything~2~\cite{ravi2024sam} for mask segmentation. DiffuEraser~\cite{li2025diffueraser} is used for inpainting. The canonical meshes of articulated objects are generated using HunYuan3D~\cite{lai2025hunyuan3d} from inpainted canonical frames.

In ASR, we run 20 iterations with an initial sampling range $\delta=0.03$. Part motion reconstruction uses 500 iterations per frame with Adam optimizer and a linearly decayed learning rate from $0.02$ to $0.002$. The loss weights are set to $\lambda_{\mathrm{match}}=1.0$ and $\lambda_{\mathrm{smooth}}=0.01$.

For articulated HOI alignment, we employ Qwen-VL-Max~\cite{Qwen-VL} for MLLM-based contact reasoning, followed by 800 optimization steps over all frames with Adam. The learning rate decreases from $10^{-3}$ to $10^{-4}$ and the loss weights are set to $\lambda_{\mathrm{contact}}=1$, $\lambda_{\mathrm{accel}}=1$, and $\lambda_{\theta}=50.0$.

\subsection{Evaluation Settings}

As no existing method reconstructs hand-articulated-object interactions from monocular RGB video without pre-scanned or template object templates, we compare against RSRD~\cite{kerr2024rsrd}, a recent 4D articulated HOI reconstruction approach that requires pre-scanned sequences of the object, and EasyHOI\cite{liu2025easyhoi}, a monocular image HOI reconstruction method by apply it frame-by-frame.

For evaluating 4D reconstruction of articulated objects, we report the Chamfer distance (CD) and the Maximum Symmetry-Aware Surface Distance (MSSD)~\cite{hodan2024bop} and F-score at 5mm and 10mm thresholds.  
For evaluating hand-object alignment, we adopt the Collision-Contact ($Co^2$) score from Open3DHOI~\cite{wen2025reconstructing} to evaluate 3D interaction quality, computing both contact and collision scores on annotated contact frames, and only the collision score on in-contact frames.

\begin{table}[t]
\centering
\caption{4D reconstruction accuracy of articulated object on monocular RGB videos from ArtHOI-RGBD dataset. Lower CD/MSSD and higher F-scores indicate better performance.}
\vspace{-2mm}
\label{tab:part_pose_compact_cr}
\resizebox{\columnwidth}{!}{%
\begin{tabular}{lcccccccc}
\toprule
Objects & Method & CD (mm)\,$\downarrow$ & MSSD (mm)\,$\downarrow$ & F10\,$\uparrow$ & F5\,$\uparrow$ \\
\midrule
\multirow{3}{*}{Headphone} 
& EasyHOI~\cite{liu2025easyhoi}       & \valuestd{209.295}{107.36} & \valuestd{291.02}{105.03} & 1.26 & 0.59 \\
& RSRD~\cite{kerr2024rsrd}        & \valuestd{14.708}{0.18} & \valuestd{41.06}{3.65} & 41.67 & 20.91 \\
& Ours          & \textbf{\valuestd{8.124}{0.44}} & \textbf{\valuestd{30.43}{2.20}} & \textbf{69.68} & \textbf{42.19} \\

\midrule
\multirow{3}{*}{Scissor} 
& EasyHOI~\cite{liu2025easyhoi}       & \valuestd{170.946}{110.29} & \valuestd{210.08}{114.00} & 4.14 & 2.24 \\
& RSRD~\cite{kerr2024rsrd}        & \valuestd{13.841}{5.89} & \valuestd{31.53}{5.20} & 37.98 & 17.55 \\
& Ours               & \textbf{\valuestd{4.256}{1.02}} & \textbf{\valuestd{15.14}{3.14}} & \textbf{92.57} & \textbf{65.00} \\
\midrule
\multirow{3}{*}{Candy Box} 
& EasyHOI~\cite{liu2025easyhoi}       & \valuestd{50.693}{55.08} & \valuestd{82.36}{67.44} & 25.96 & 12.94 \\
& RSRD~\cite{kerr2024rsrd}        & \valuestd{7.768}{4.88} & \valuestd{31.40}{25.10} & 83.11 & 55.25\\
& Ours               & \textbf{\valuestd{4.104}{1.33}} & \textbf{\valuestd{17.67}{5.90}} & \textbf{92.55} & \textbf{71.63} \\
\midrule
\multirow{3}{*}{CD Drive} 
& EasyHOI~\cite{liu2025easyhoi}       & \valuestd{648.704}{391.84} & \valuestd{827.14}{400.53} & 0.08 & 0.04 \\
& RSRD~\cite{kerr2024rsrd}        & \valuestd{282.330}{192.30} & \valuestd{348.59}{233.49} & 10.90 & 6.92 \\
& Ours               & \textbf{\valuestd{3.334}{0.20}} & \textbf{\valuestd{9.71}{1.89}} & \textbf{96.01} & \textbf{78.75} \\
\midrule
\multirow{3}{*}{Stapler} 
& EasyHOI~\cite{liu2025easyhoi}       & \valuestd{116.813}{96.20} & \valuestd{198.25}{97.52} & 8.58 & 5.28 \\
& RSRD~\cite{kerr2024rsrd}        & \valuestd{288.704}{144.44} & \valuestd{363.92}{106.98} & 0.80 & 0.34 \\
& Ours               & \textbf{\valuestd{4.487}{0.97}} & \textbf{\valuestd{20.15}{5.81}} & \textbf{91.63} & \textbf{67.94} \\
\bottomrule
\end{tabular}%
}
\vspace{-2mm}
\end{table}

\begin{table}[t]
\centering
\caption{4D reconstruction accuracy of articulated object on monocular RGB videos from RSRD~\cite{kerr2024rsrd} dataset. Lower CD/MSSD and higher F-scores indicate better performance.}
\vspace{-2mm}
\label{tab:part_pose_rsrdrgb_compact_cr}
\resizebox{\columnwidth}{!}{%
\begin{tabular}{lcccccccc}
\toprule
Objects & Method & CD (mm)\,$\downarrow$ & MSSD (mm)\,$\downarrow$ & F10\,$\uparrow$ & F5\,$\uparrow$ \\
\midrule
\multirow{3}{*}{Scissor} 
& EasyHOI~\cite{liu2025easyhoi}       & \valuestd{205.567}{214.40} & \valuestd{270.99}{232.23} & 6.04 & 3.07 \\
& RSRD~\cite{kerr2024rsrd}        & \valuestd{68.564}{20.94} & \valuestd{109.31}{19.91} & 2.09 & 1.20 \\
& Ours          & \textbf{\valuestd{5.447}{3.34}} & \textbf{\valuestd{13.12}{6.38}} & \textbf{80.95} & \textbf{61.81} \\
\midrule
\multirow{3}{*}{LED Light} 
& EasyHOI~\cite{liu2025easyhoi}       & \valuestd{52.435}{61.43} & \valuestd{100.18}{72.43} & 16.43 & 7.64 \\
& RSRD~\cite{kerr2024rsrd}        & \textbf{\valuestd{10.144}{1.31}} & \textbf{\valuestd{32.80}{5.72}} & \textbf{63.21} & \textbf{46.21} \\
& Ours          & \valuestd{10.836}{7.50} & \valuestd{35.30}{21.84} & 60.95 & 31.81 \\
\midrule
\multirow{3}{*}{Bear} 
& EasyHOI~\cite{liu2025easyhoi}       & \valuestd{23.785}{2.85} & \valuestd{77.29}{14.88} & 27.45 & 12.89 \\
& RSRD~\cite{kerr2024rsrd}        & \textbf{\valuestd{8.739}{1.03}} & \textbf{\valuestd{28.70}{3.63}} & \textbf{65.63} & \textbf{32.73} \\
& Ours          & \valuestd{12.374}{2.34} & \valuestd{30.31}{5.76} & 45.48 & 22.86 \\
\midrule
\multirow{3}{*}{Sunglasses} 
& EasyHOI~\cite{liu2025easyhoi}       & \valuestd{123.385}{56.65} & \valuestd{304.35}{84.76} & 3.13 & 1.61 \\
& RSRD~\cite{kerr2024rsrd}        & \valuestd{31.985}{8.46} & \valuestd{164.60}{63.75} & 28.44 & 15.66 \\
& Ours          & \textbf{\valuestd{9.956}{4.38}} & \textbf{\valuestd{41.38}{20.84}} & \textbf{65.14} & \textbf{44.39} \\
\bottomrule
\end{tabular}%
}
\vspace{-2mm}
\end{table}

\begin{table}[t]
\centering
\small
\setlength{\tabcolsep}{6pt}
\caption{Comparison on a subset of ARCTIC~\cite{fan2023arctic}. `Cont.Acc' denotes binary contact accuracy and `Fing.Acc' denotes main contacting finger (thumb, index, middle) accuracy of MLLM reasoning results.}
\vspace{-2mm}
\label{tab:rebuttal_ARCTIC}
\resizebox{\linewidth}{!}{
\begin{tabular}{llccccc}
\hline
 & Method & CD(mm) $\downarrow$ & F10 $\uparrow$ & MSSD(mm) $\downarrow$ & Cont.Acc\% $\uparrow$ & Fing.Acc\% $\uparrow$ \\
\hline

\multirow{3}{*}{Mixer}
 & EasyHOI~\cite{liu2025easyhoi} & 226.0 & 0.05 & 326.1 & N/A & N/A\\
 & RSRD~\cite{kerr2024rsrd}    & \multicolumn{5}{c}{Failed to reconstruct articulated object} \\
 & Ours    & \textbf{12.1} & \textbf{0.55} & \textbf{41.4} & \textbf{82.5} &  \textbf{76.2} \\

\hline

\multirow{3}{*}{Box}
 & EasyHOI~\cite{liu2025easyhoi} & 207.9 & 0.03 & 356.4 & N/A & N/A \\
 & RSRD~\cite{kerr2024rsrd}    & \multicolumn{5}{c}{Failed to reconstruct articulated object} \\
 & Ours    & \textbf{14.0} & \textbf{0.44} & \textbf{51.2} & \textbf{76.6} & \textbf{62.3}  \\

\hline

\multirow{3}{*}{Scissors}
 & EasyHOI~\cite{liu2025easyhoi} & 436.4     & 0.01     & 497.1 & N/A & N/A \\
 & RSRD~\cite{kerr2024rsrd}    & \multicolumn{5}{c}{Failed to reconstruct articulated object} \\
 & Ours    & \textbf{58.6} & \textbf{0.10} & \textbf{185.2} & \textbf{57.9} & \textbf{54.4} \\

\hline
\end{tabular}
}
\vspace{-5mm}
\end{table}

\begin{table}[t]
\centering
\caption{Comparison of $Co^2$ scores for unaligned and aligned articulated HOI reconstruction under different contact reasoning strategies. We evaluate four settings: (1) unaligned hand-object reconstruction, (2) RSRD with WiLoR~\cite{potamias2025wilor} hands, (3) our alignment using a mask-intersection contact heuristic (w/o MLLM), and (4) our full alignment with MLLM-based contact reasoning (w/ MLLM). Lower is better. RSRD fails on ArtHOI-Wild due to missing object-scanning inputs.}
\vspace{-3mm}
\label{tab:vlm_contact}
\resizebox{\columnwidth}{!}{%
\begin{tabular}{lccc}
\toprule
 & 
ArtHOI-RGBD & 
RSRD~\cite{kerr2024rsrd} & 
ArtHOI-Wild \\
\midrule
No Alignment  & 0.972 & 0.517 & 0.514 \\
RSRD~\cite{kerr2024rsrd} + WiLoR~\cite{potamias2025wilor}         &  0.392 & 0.166 & N/A \\
Ours w/o MLLM     & 0.046 & 0.035 & 0.059  \\
Ours w/ MLLM     & \textbf{0.029} & \textbf{0.022} & \textbf{0.039}  \\
\bottomrule
\end{tabular}%
}
\vspace{-6mm}
\end{table}

\subsection{Quantitative Results}

We evaluate our method on three aspects: the accuracy of articulated object reconstruction, the quality of overall hand-object interaction (HOI) alignment and the accuracy of MLLM-driven contact reasoning results.

\noindent\textbf{Articulated Object Reconstruction Quality.}  
We evaluate articulated object 4D reconstruction on annotated sequences from ArtHOI-RGBD, RSRD and ARCTIC~\cite{fan2023arctic}. For a fair comparison, 3D Gaussian part representation of RSRD is replaced with the corresponding mesh during evaluation. Tables~\ref{tab:part_pose_compact_cr}, \ref{tab:part_pose_rsrdrgb_compact_cr} and \ref{tab:rebuttal_ARCTIC} shows that, on all five ArtHOI-RGBD sequences featuring challenging hand-part occlusions (e.g., Stapler) and part-part occlusions (e.g., CD Drive), our method achieves consistently lowest reconstruction errors. On the RSRD dataset, our results are comparable to RSRD despite not requiring any pre-scanning. In addition, our approach successfully handles ArtHOI-Wild and ARCTIC videos, whereas RSRD fails due to the absence of a surrounding scan.

\noindent\textbf{HOI Alignment Quality.}  
We assess the final HOI alignment using the collision-contact ($Co^2$) score. Table~\ref{tab:vlm_contact} and Fig.~\ref{fig:fig_hoalign} compare unaligned outputs, RSRD (with WiLoR~\cite{potamias2025wilor} hand estimates) and our MLLM-guided alignment. Our optimization, guided by MLLM-derived contact cues, produces the lowest $Co^2$ scores and visually plausible, well-aligned 4D reconstructions, outperforming competing strategies that lack scale-aware or temporally consistent contact constraints.

\noindent\textbf{MLLM Contact Reasoning Accuracy.}  
Table~\ref{tab:vlm_ablation} reports contact accuracy and the false-positive (FP) rates. To account for temporal ambiguity at interaction boundaries, predictions within $\pm 1\!-\!3$ frames of the annotated contact window are counted as correct.The results show that our prompting scheme substantially reduces FP while improving accuracy, particularly on in-the-wild data.

\subsection{Qualitative Results}

Fig.~\ref{fig:fig_hoalign} presents qualitative comparisons across all datasets. Our method robustly reconstructs articulated object geometry and motion, together with aligned interacting hands in both controlled and in-the-wild scenarios, demonstrating strong robustness and practical applicability in real-world settings. In-the-wild sequences often exhibit substantial occlusions between hand-part and part-part, making reconstruction particularly challenging. Even under such conditions, our framework maintains coherent geometry and motion across frames by leveraging consistent geometric, depth, and interaction cues extracted from diverse foundation models. In contrast, RSRD struggles with heavy occlusions and latent ambiguities and fails to produce precise part-motion trajectories. Importantly, our method generalizes robustly to in-the-wild videos, successfully recovering both part motion and hand alignment. In contrast, RSRD and other similar approaches require a pre-scanned object in a canonical state, which is infeasible for internet videos and often unattainable even for lab-captured interaction videos.

\begin{table}[t]
\centering
\caption{Comparison of canonical mesh pose and scale optimization. We compare with FoundationPose and Any6D~\cite{lee2025any6d}. Metrics include the IoU between rendered and ground-truth masks under the optimized pose, and the optimization success rate (SR$\%$). A case is considered failed if subsequent part motion reconstruction or HOI alignment cannot proceed.}
\vspace{-3mm}
\small
\label{tab:abl_canonical_reg}
\resizebox{\columnwidth}{!}{%
\begin{tabular}{lccccccc}
\toprule
\multirow{2}{*}{Method} &
\multicolumn{2}{c}{ArtHOI-RGBD} &
\multicolumn{2}{c}{RSRD~\cite{kerr2024rsrd}} &
\multicolumn{2}{c}{ArtHOI-Wild} \\
\cmidrule(lr){2-3}
\cmidrule(lr){4-5}
\cmidrule(lr){6-7}
& IoU$\uparrow$ & SR($\%$)$\uparrow$
& IoU$\uparrow$ & SR($\%$)$\uparrow$
& IoU$\uparrow$ & SR($\%$)$\uparrow$ \\
\midrule
FoundationPose~\cite{wen2024foundationpose} &  0.820 & 60\% & 0.706 & 78\% & 0.749 & 71\% \\
Any6D~\cite{lee2025any6d}          &  0.876 & 60\% & 0.857 & 78\% & 0.683 & 57\% \\
ASR (ours)     &  \textbf{0.905} & \textbf{100\%} & \textbf{0.876} & \textbf{100\%} & \textbf{0.882} & \textbf{100\%} \\
\bottomrule
\end{tabular}%
}
\vspace{-3mm}
\end{table}

\begin{table}[t]
\centering
\caption{Ablation study on prompting strategies for MLLM contact reasoning, evaluated by accuracy and false positive rate (FP, $\%$). ``Temp.'' incorporates temporal context from neighboring frames. ``Persp.'' indicates introducing camera-perspective cues; ``MinFP'' uses prompts designed to suppress false positives; and ``Depth'' augments image prompts with colorized depth. Results of ArtHOI-RGBD is excluded due to its near $100\%$ accuracy.}
\vspace{-3mm}
\label{tab:vlm_ablation}
\resizebox{\columnwidth}{!}{%
\begin{tabular}{cccccccc}
\toprule
\multicolumn{4}{c}{Prompting Strategy} 
& \multicolumn{2}{c}{RSRD~\cite{kerr2024rsrd}} 
& \multicolumn{2}{c}{ArtHOI-Wild} \\
\cmidrule(lr){1-4}
\cmidrule(lr){5-6}
\cmidrule(lr){7-8}
Temp. & Persp. & MinFP & Depth 
& Acc.\,$\uparrow$ & FP\,$\downarrow$ 
& Acc.\,$\uparrow$ & FP\,$\downarrow$ \\
\midrule
            &            &            &            &  81.53 & 18.24 & 83.50 & 16.59 \\
\checkmark  &            &            &            &  82.75 & 17.08 & 82.62 & 17.02 \\
\checkmark  &            & \checkmark & \checkmark &  86.42 & 13.49 & 85.92 & 13.27 \\
\checkmark  & \checkmark &            & \checkmark &  86.27 & 13.66 & 86.21 & 13.79 \\
\checkmark  & \checkmark & \checkmark &            &  \underline{87.65} & \underline{12.13} & \textbf{87.52} & \underline{11.35} \\
\checkmark  & \checkmark & \checkmark & \checkmark &  \textbf{88.58} & \textbf{11.20} & \underline{86.56} & \textbf{9.81} \\
\bottomrule
\end{tabular}%
}
\vspace{-4mm}
\end{table}

\subsection{Ablation Study}
\label{sec:ablation}

\begin{figure}
  \centering
  \includegraphics[width=0.99\linewidth]{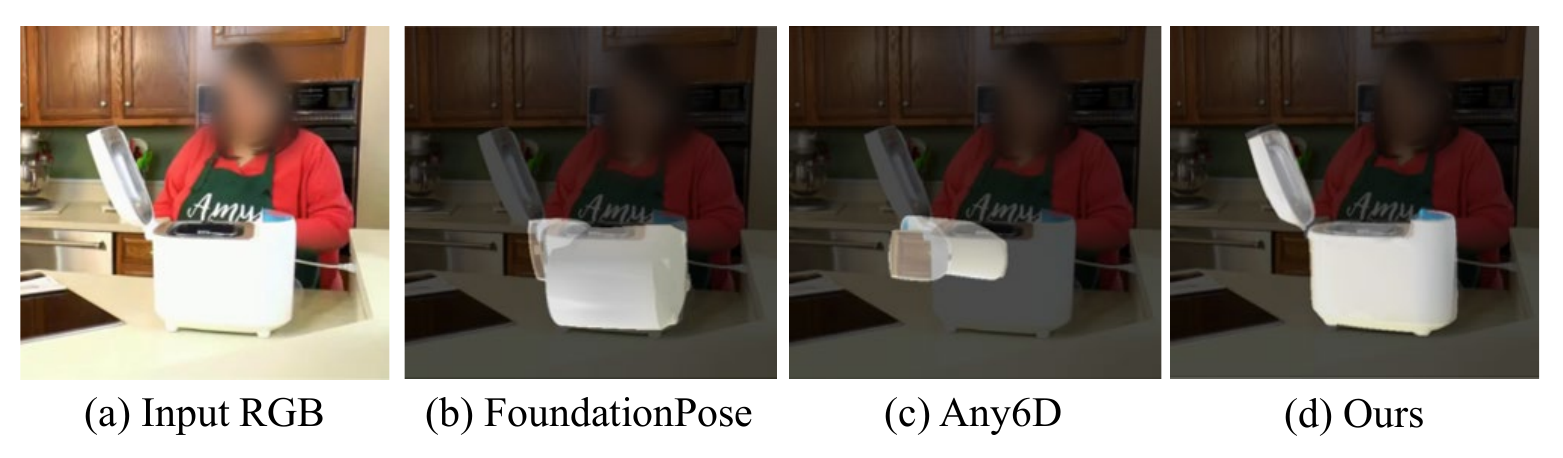}
  \vspace{-2mm}
  \caption{Qualitative comparison of metric scale and pose estimation on in-the-wild videos without ground-truth depth. Images are cropped and zoomed-in for better visualization.}
  \label{fig:fig_abl_asr}
\vspace{-5mm}
\end{figure}

\noindent\textbf{Effect of Adaptive Sampling Refinement.}
We evaluate the effectiveness of Adaptive Sampling Refinement (ASR) by comparing it against directly applying FoundationPose~\cite{wen2024foundationpose} using only the coarse scale estimate. We further include Any6D~\cite{lee2025any6d}, a model-free RGB-D method for scale and 6-DoF pose estimation, as it follows a conceptually similar strategy and can be adapted to our setting. For a fair comparison with Any6D, we use the same HunYuan3D mesh and match the number of scale samples used in ASR. Table~\ref{tab:abl_canonical_reg} reports the 2D silhouette IoU and optimization success rate, where a failure is defined as any case in which subsequent part-motion reconstruction or HOI alignment cannot proceed. ASR achieves the highest IoU and success rates across all videos. In contrast, FoundationPose often fails due to inconsistencies between the generated mesh and noisy depth estimates, while Any6D struggles to recover a valid metric scale owing to its dependence on empirically tuned hyperparameters. Fig.~\ref{fig:fig_abl_asr} provides a qualitative comparison on an ArtHOI-Wild example.

\noindent\textbf{Effect of MLLM-guided Hand-Object Interaction Alignment.}
We ablate the proposed MLLM-guided HOI alignment by comparing against three variants: (1) a baseline that removes the alignment module entirely, (2) RSRD hand-object reconstruction using WiLoR as the hand estimator, and (3) a simple heuristic that infers contact from hand-object mask intersection. As shown in Table~\ref{tab:vlm_contact}, excluding MLLM-derived contact cues consistently degrades reconstruction accuracy. Qualitative results in Fig.~\ref{fig:fig_hoalign} further highlight that, without scale and spatio-temporal optimization on hand and object parameters, the reconstructed 4D hand and articulated object suffer from severe spatial drift and scale inconsistency, revealing the necessity of MLLM-guided HOI alignment.

\noindent\textbf{Effect of Prompting Strategies in MLLM Reasoning.}
We ablate the effect of four prompting components: temporal context (Temp.), camera perspective cues (Persp.), false positive suppression (MinFP), and depth-augmented image prompts (Depth) by progressively enabling them and reporting accuracy and FP in Table~\ref{tab:vlm_ablation}. Incorporating temporal context provides modest gains, while adding perspective reasoning and MinFP prompts substantially reduces spurious contact predictions. Temporal and depth-augmented prompts further improve robustness on challenging in-the-wild videos where single-frame appearance cues are unreliable. The full combination of all components produces the best trade-off, achieving the highest accuracy and lowest FP across both RSRD and ArtHOI-Wild.

\section{Conclusion}
We presented a method for reconstructing 4D hand-object interactions with articulated objects from monocular videos. Our approach leverages rich priors from multiple foundation models and unifies them through optimization strategies that explicitly handle cross-prior inconsistencies and estimation noise. Extensive experiments on two datasets demonstrate that our model-free method outperforms prior approaches relying on pre-scanned articulated objects, and generalizes effectively to in-the-wild Internet videos, showcasing robust real-world applicability to articulated interactions.

\section*{Acknowledgement}

This work was partially supported by the National Key RD Program of China under Grant No. 2022YFA1004100 and China Postdoctoral Science Foundation under Grant No. 2025M784371.

{
    \small
    \bibliographystyle{ieeenat_fullname}
    \bibliography{main}
}

\clearpage
\setcounter{page}{1}
\maketitlesupplementary

\renewcommand{\thesection}{\Alph{section}}
\renewcommand{\thetable}{\Alph{table}}
\renewcommand{\thefigure}{\Alph{figure}}
\setcounter{section}{0}
\setcounter{figure}{0}
\setcounter{table}{0}

\begin{figure}[t]
  \centering
  \includegraphics[width=0.99\linewidth]{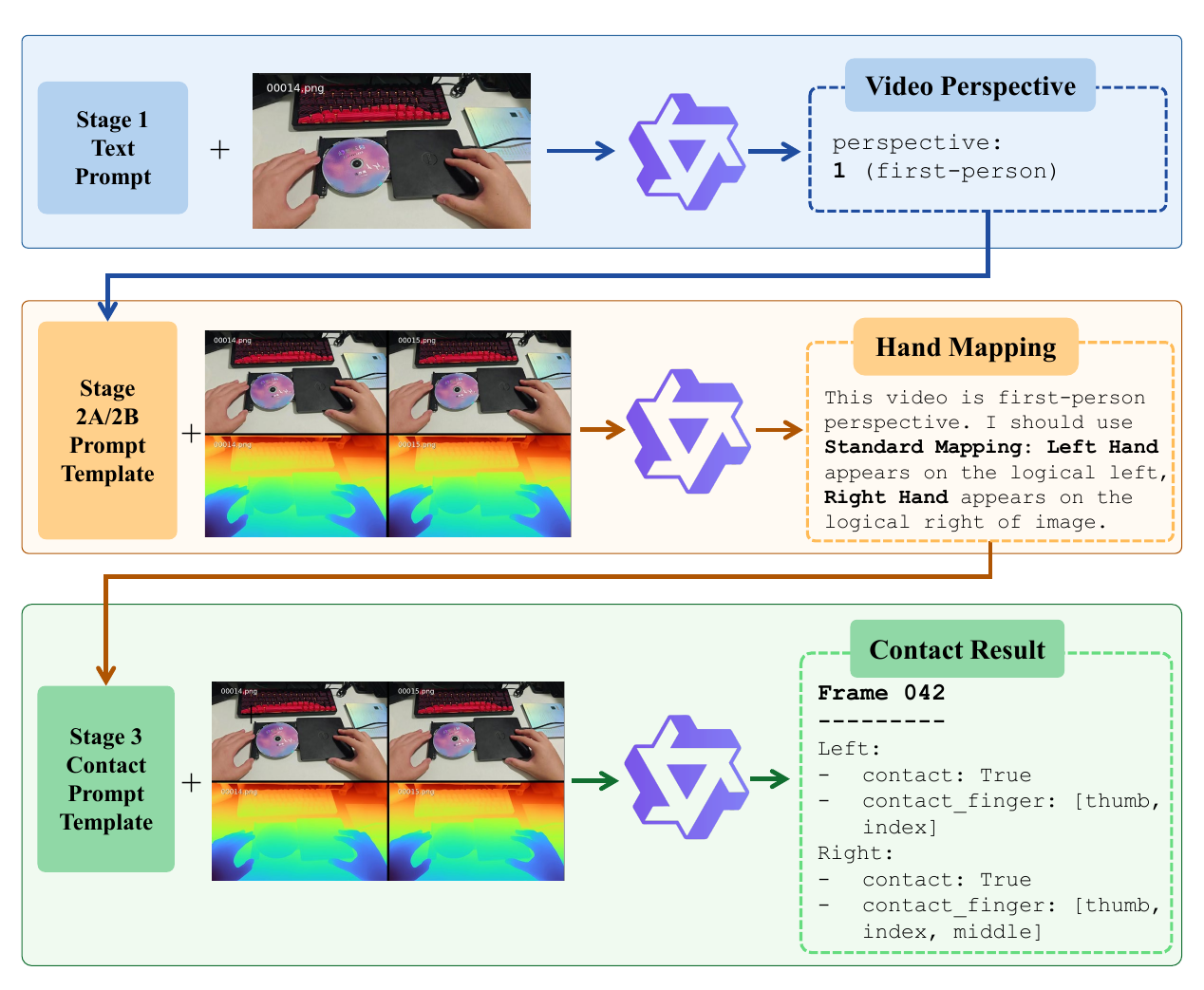}
  \caption{Demonstration of our MLLM contact reasoning pipeline. For clarity, we merge 2 neighbouring frames, but in practice, it's typically set to 3. The top row shows RGB frames, the bottom row shows colorized depth maps. The MLLM analyzes visual and depth cues across frames to determine contact status and engaged fingers for each hand.}
  \label{fig:fig_mllm_demo}
\end{figure}

\section{Implementation Details}

\subsection{Coarse Metric Scale Estimation of Object}

We detail the coarse scale estimation introduced in Sec. 3.2. Given estimated metric depth maps, we first back-project them into 3D space using camera intrinsics $\mathbf{K}$ and the object mask. To suppress boundary noise, the mask is eroded prior to back-projection, followed by a Statistical Outlier Removal (SOR) filter to further clean the point cloud. We then compute the bounding boxes of both the normalized canonical object and the back-projected depth point cloud. The coarse metric scale $s_{\mathrm{coarse}}^o$ is obtained as the maximum ratio between their extents along the x- and y-axes. The z-axis (depth direction) is excluded because the back-projected point cloud only captures the visible object surface and is typically more noisy and unreliable in depth.

\begin{table}[t]
\centering
\caption{Comparison of contact accuracy (Acc.) and false positive rate (FP) between our MLLM-based contact reasoning and a rule-based mask-intersection heuristic. While both methods perform similarly on the controlled RSRD dataset, the heuristic degrades notably on in-the-wild videos, whereas the MLLM remains robust. ArtHOI-RGBD is excluded due to its near-perfect accuracy.}
\label{tab:mllmvsmask}
\begin{tabular}{lcccc}
\toprule
& \multicolumn{2}{c}{RSRD~\cite{kerr2024rsrd}} & \multicolumn{2}{c}{ArtHOI-Wild} \\
\cmidrule(lr){2-3} \cmidrule(lr){4-5}
Contact Judge & Acc. $\uparrow$ & FP $\downarrow$ & Acc. $\uparrow$ & FP $\downarrow$ \\
\midrule
Mask Intersection & 0.86 & 0.14 & 0.76 & 0.23 \\
MLLM & \textbf{0.89} & \textbf{0.11} & \textbf{0.87} & \textbf{0.10} \\
\bottomrule
\end{tabular}
\end{table}

\begin{figure*}[t]
  \centering
  \includegraphics[width=0.99\linewidth]{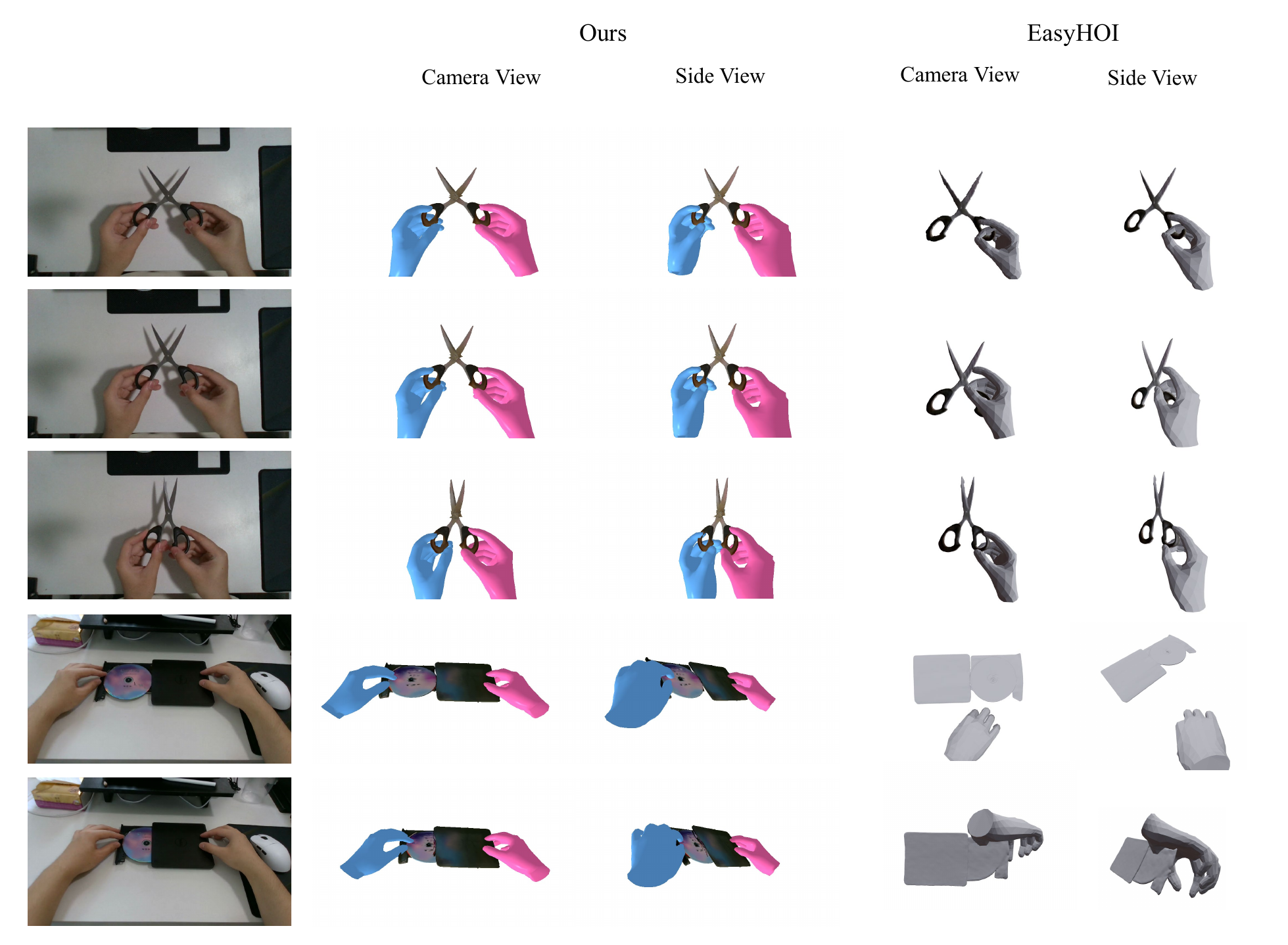}
  \caption{Qualitative comparison between our method and EasyHOI~\cite{liu2025easyhoi} on ArtHOI-RGBD. EasyHOI often fails to recover articulated object scale and pose, and exhibits inconsistent hand-object alignment across frames.}
  \label{fig:fig_ours_easyhoi}
\end{figure*}

\subsection{Object Part Segmentation}

Sec. 3.3 describes the reconstruction of part-wise motion for articulated objects. Here, we provide additional details on the part partition process. We begin by applying PartField~\cite{liu2025partfield} to extract per-vertex feature fields, followed by agglomerative clustering to obtain vertex group labels. The object is then rendered in its canonical pose using PyTorch3D~\cite{ravi2020pytorch3d} to produce a 2D label map. Vertex groups are merged according to part masks, after which the mesh is finally split into individual parts.

\subsection{MLLM Contact Reasoning}

We adopt an image-text question-answer strategy to extract contact information for each frame of input video. The primary challenge of this task lies in suppressing false positives: in real-world videos, both humans and models often confuse near-contact with genuine physical contact, while clear separation is seldom misidentified as contact, making false negatives comparatively rare. To mitigate this, we augment RGB frames with colorized depth, incorporate neighboring-frame sampling to strengthen spatio-temporal cues, and explicitly instruct the MLLM to be cautious about false positives. Furthermore, because the input videos may be egocentric or exocentric, we identify video perspective beforehand to reduce hallucinations on hand laterality when reasoning about bimanual contact. Figures~\ref{fig:mllm_prompts_s1}, \ref{fig:mllm_prompts_s2}, and \ref{fig:mllm_prompts_s3} demonstrate the full prompt templates used in our pipeline.

\paragraph{Input and Output Format}

To provide richer contextual cues, we concatenate $k$ neighboring frames ($k=3$ in practice) along with their colorized depth maps into a single large image prompt, which the MLLM can jointly analyze for spatio-temporal consistency. The depth maps are visualized with a color gradient (blue for near, red for far), making depth discontinuities visually salient to the model. The output is a structured JSON containing: (i) frame count and which hands appeared in the video; (ii) for each frame, binary contact flags for left and right hands; (iii) lists of contacting fingers for each hand-frame pair, empty if no contact. This structured format enables downstream optimization to directly parse and apply contact constraints.

\paragraph{Three-Stage Prompting Strategy}

The MLLM contact reasoning pipeline consists of three carefully designed stages, as shown in Figures~\ref{fig:mllm_prompts_s1}, \ref{fig:mllm_prompts_s2}, and \ref{fig:mllm_prompts_s3}.

Stage 1: Perspective Detection. Video perspective (egocentric vs. exocentric) significantly affects hand laterality interpretation. In first-person perspective, a single visible hand is automatically from the operator's viewpoint, and spatial relationships are relatively straightforward. In third-person perspective, MLLM must infer the operator's orientation and account for mirror effects to correctly identify hands. By first explicitly determining the perspective, we reduce hallucinations on hand identity in subsequent reasoning stages.

Stage 2: Hand Mapping. After identifying perspective, hand mapping disambiguates left and right hands through perspective-specific heuristics. For first-person videos (Stage 2a), spatial positioning and thumb direction provide direct cues. For third-person videos (Stage 2b), the strategy shifts to analyzing relationship between camera and the operator's body. In this stage, the MLLM can map visible hands to left or right labels.

Stage 3: Frame-wise Contact Reasoning. Given correct hand identity, Stage 3 performs detailed frame-by-frame contact analysis. For each visible hand, the prompt guides the MLLM through a structured reasoning chain. The prompt emphasizes caution: uncertain cases should be marked as no-contact ($\texttt{false}$) to suppress false positives. This conservative bias aligns with our observation that false positive predictions in real-world contact cases are more often than false negatives.


\definecolor{prompt_bg_A}{RGB}{236,244,255} 
\definecolor{prompt_bg_B}{RGB}{255,245,230} 
\definecolor{prompt_bg_C}{RGB}{237,251,240} 
\definecolor{prompt_title_A}{RGB}{31,78,160}
\definecolor{prompt_title_B}{RGB}{175,84,0}
\definecolor{prompt_title_C}{RGB}{14,108,46}

\newcommand{\prompttitle}[2]{\textbf{\textcolor{#1}{#2}}}

\begin{figure*}[t]
  \centering
  \begin{tcolorbox}[
    enhanced, breakable,
    colback=prompt_bg_A, colframe=prompt_title_A,
    boxrule=0.7pt, arc=3pt,
    left=4pt, right=4pt, top=3pt, bottom=3pt,
    fontupper=\fontsize{6.3pt}{8pt}\selectfont\ttfamily,
    title={\prompttitle{white}{Stage 1 ~~Perspective Detection Prompt ~~(\texttt{prompt\_perspective})}},
    coltitle=white, attach boxed title to top left={yshift=-2mm, xshift=4mm},
    boxed title style={colback=prompt_title_A, arc=2pt, boxrule=0pt},
    width=\linewidth
  ]
You are given a set of images sampled from a video about a human manipulating an articulated object.\par
Please determine whether this video is from a \textbf{first-person perspective} or a \textbf{third-person perspective}.\par\smallskip
If it is first-person perspective, output only the number \texttt{1}.\par
If it is third-person perspective, output only the number \texttt{3}.\par
Do not output any other text, explanation, or thoughts.\par\smallskip
\textbf{First-person}: filmed from the operator's point of view; arms/hands extend from the bottom or sides of the frame; viewpoint aligned with the operator's head direction.\par
\textbf{Third-person}: filmed from an observer's point of view; shows the whole/most of the operator's body; viewpoint not aligned with the operator's head direction.\par\smallskip
\textit{Judgment principles:}\par
\quad 1. If only one hand appears, it must be first-person perspective.\par
\quad 2. If a human face appears, it must be third-person perspective.\par
\quad 3. In first-person perspective, the hand(s) usually occupy a large area of the image.
  \end{tcolorbox}

  \caption{Stage 1: Perspective Detection Prompt.
  This prompt determines whether the input video is from a first-person or third-person viewpoint, which is essential for correctly identifying hand laterality in subsequent stages.}
  \label{fig:mllm_prompts_s1}
\end{figure*}

\begin{figure*}[t]
  \centering
  \begin{minipage}[t]{0.49\linewidth}
  \begin{tcolorbox}[
    enhanced, breakable,
    colback=prompt_bg_B, colframe=prompt_title_B,
    boxrule=0.7pt, arc=3pt,
    left=4pt, right=4pt, top=3pt, bottom=3pt,
    fontupper=\fontsize{6.3pt}{8pt}\selectfont\ttfamily,
    title={\prompttitle{white}{Stage 2a ~~Hand Mapping — First-Person}},
    coltitle=white, attach boxed title to top left={yshift=-2mm, xshift=4mm},
    boxed title style={colback=prompt_title_B, arc=2pt, boxrule=0pt},
    height from=0pt to 999pt
  ]
\textbf{Step 1: Perspective Analysis \& Hand Mapping (Reasoning Chain of Thought Example)}\par
This video is \textbf{First-person} perspective. The camera mimics the operator's eyes. I need to determine carefully about hand side.\par
1. If there's two hands, then the hand on the left side of the image is the Left Hand, and the hand on the right side is the Right Hand. If only one hand appears, I need to determine carefully.\par
1. \textbf{Standard Mapping}: Usually, the Left Hand enters from the logical left, and the Right Hand from the logical right.\par
2. But I also need to examine the thumb direction to determine. Thumb of left hand is pointing right, and thumb of right hand is pointing left.\par
So I can check the thumb direction to determine the hand side, especially when only one hand appears.\par
3. \textbf{Apply Mapping}: Use these cues to strictly confirm the identity of any visible hands before proceeding.
  \end{tcolorbox}
  \end{minipage}
  \hfill
  \begin{minipage}[t]{0.49\linewidth}
  \begin{tcolorbox}[
    enhanced, breakable,
    colback=prompt_bg_B, colframe=prompt_title_B,
    boxrule=0.7pt, arc=3pt,
    left=4pt, right=4pt, top=3pt, bottom=3pt,
    fontupper=\fontsize{6.3pt}{8pt}\selectfont\ttfamily,
    title={\prompttitle{white}{Stage 2b ~~Hand Mapping — Third-Person}},
    coltitle=white, attach boxed title to top left={yshift=-2mm, xshift=4mm},
    boxed title style={colback=prompt_title_B, arc=2pt, boxrule=0pt},
    height from=0pt to 999pt
  ]
\textbf{Step 1: Perspective Analysis \& Hand Mapping (Chain of Thought)}\par
This video is \textbf{Third-person} perspective. You must determine the exact camera angle to identify hands correctly:\par
1. \textbf{Analyze Operator Orientation}: Look at the person's body/head in the RGB frames.\par
2. \textbf{Determine View Type \& Arm Connectivity}:\par
\quad -- \textbf{Frontal/Side-Front View}: Camera faces the person. $\to$ \textbf{Logic}: Mirror effect (Left Hand is on Right, Right Hand is on Left).\par
\quad -- \textbf{Rear/Side-Rear View}: Camera looks at the person's back or side-back. Use \textbf{Arm Connectivity} to identify hands:\par
\qquad * \textbf{Right Side-Rear}: Camera observes from the operator's right-back. The \textbf{Right Arm} is visibly connected to the body on the right. $\to$ \textbf{Logic}: The hand connected to this visible right arm is the \textbf{Right Hand}. The other hand is the Left Hand.\par
\qquad * \textbf{Left Side-Rear}: Camera observes from the operator's left-back. The \textbf{Left Arm} is visibly connected to the body on the left. $\to$ \textbf{Logic}: The hand connected to this visible left arm is the \textbf{Left Hand}. The other hand is the Right Hand.\par
3. \textbf{Apply Mapping}: Based on the arm connections, strictly assign 'Left Hand' and 'Right Hand' labels before proceeding.
  \end{tcolorbox}
  \end{minipage}

  \caption{Stage 2: Hand Mapping Prompt.
  This stage identifies and maps visible hands to left/right labels. Stage 2a handles first-person perspective videos using spatial positioning and thumb direction cues. Stage 2b handles third-person perspective videos by analyzing camera angle relative to the operator's body and arm connectivity patterns.}
  \label{fig:mllm_prompts_s2}
\end{figure*}

\begin{figure*}[t]
  \centering

  \begin{tcolorbox}[
    enhanced, breakable,
    colback=prompt_bg_C, colframe=prompt_title_C,
    boxrule=0.7pt, arc=3pt,
    left=4pt, right=4pt, top=3pt, bottom=3pt,
    fontupper=\fontsize{6.3pt}{8pt}\selectfont\ttfamily,
    title={\prompttitle{white}{Stage 3 ~~Contact Reasoning Prompt ~~(\texttt{prompt}, appended after Stage 2)}},
    coltitle=white, attach boxed title to top left={yshift=-2mm, xshift=4mm},
    boxed title style={colback=prompt_title_C, arc=2pt, boxrule=0pt},
    width=\linewidth
  ]
\textbf{Step 2: Frame-by-Frame Contact Reasoning}\par
The image contains K frames (horizontally merged) separated by black bars.\par
\quad -- Top row: \textbf{RGB frames.}\quad -- Bottom row: \textbf{Depth frames} (\textcolor{blue}{blue}=near, \textcolor{red}{red}=far).\par\smallskip
For each frame, analyze the 'Left Hand' and 'Right Hand' (identified in Step 1) separately using the following Chain of Thought:\par
\quad A. \textbf{Visibility Check}: Is the hand visible? If not, skip.\par
\qquad (A.1) Left or Right hand may be occluded by the articulated object. I need to identify partial, occluded hand around object, not missing them.\par
\quad B. \textbf{Object contact estimation}: Is the hand close enough to contact the articulated object (but not background) in the RGB frame?\par
\qquad -- If the hand is clearly distant from the object or merely contact, mark \texttt{contact:false} and skip to next.\par
\qquad (B.1) I need to carefully identify the hand appearance, fully utilizing both RGB and depth.\par
\qquad (B.2) I need to carefully determine if the hand is contacting the articulated object in a solid state, or it's in mere contact (which I should output FALSE).\par
\quad C. \textbf{Depth Map Verification (Critical Phase)}:\par
\qquad -- Look at the bottom row (Depth map) corresponding to the hand's position.\par
\qquad -- Does the hand's depth color \emph{seamlessly merge} with the object's at the interaction point?\par
\qquad -- Is there a sharp edge or color contrast separating them? If YES -> FALSE CONTACT.\par
\qquad -- Mark \texttt{contact:~true} only if depth values merge \emph{without} discontinuity.\par
\quad D. \textbf{Finger Analysis}:\par
\qquad -- If \texttt{contact:~true}, identify specific fingers (\texttt{thumb}, \texttt{index}, \texttt{middle}) involved.\par
\qquad -- If a finger is occluded or ambiguous, exclude it.\par\smallskip
\textbf{Step 3: Consistency \& Final Decision}\par
\quad -- Review your frame-by-frame findings.\par
\quad -- Ensure the contact status transitions logically (e.g., hand approaches -> touches -> leaves) and is fully supported by the visual evidence in each frame independently.\par
\quad -- Combining the neighbouring frames, make sure judge merely contact frame as FALSE contact.\par\smallskip
\textbf{Step 4: Output Generation}\par
\textbf{\textcolor{red}{IMPORTANT}}: If not 100\% sure about contact $\Rightarrow$ output \texttt{false}.\par
Do not simply output the same status for all frames; look for changes.\par
Please only output the JSON without any additional text or markdown format. output it just using text.\par\smallskip
Output format (JSON only; \texttt{r\_/l\_fingers}: valid fingers only, empty list if no contact):\par\smallskip
\begin{verbatim}
{ "frames_cnt": K,
  "appeared": ["left", "right"],  // list only hands that appeared
  "contacts": [
    { "frame": <frame_number noted on the corner of each frame, output the number using int, without .png or output string>,
      "r_contact": <bool>, "l_contact": <bool>,
      "r_fingers": ["thumb","index","middle"], // list valid fingers only, empty if no contact
      "l_fingers": [] },
    ...
  ]
}
\end{verbatim}
  \end{tcolorbox}

  \caption{Stage 3: Frame-wise Contact Reasoning Prompt.
  This stage performs detailed analysis of each frame to determine contact state and identify engaged fingers. The critical depth map verification step (Phase C) distinguishes true physical contact from mere proximity using depth discontinuity analysis.}
  \label{fig:mllm_prompts_s3}
\end{figure*}

\section{Computational Performance}

For a video sequence of 150 frames at a resolution of $960\times540$, preprocessing (mask segmentation, metric depth estimation, frame inpainting, hand estimation, and mesh reconstruction with HunYuan3D~\cite{lai2025hunyuan3d}) requires approximately 10 to 15 minutes. Optimizing the canonical object metric scale and pose takes less than 2 minutes. Part-wise motion recovery is the most time-consuming stage and takes roughly 30 minutes; during this stage, our pipeline could concurrently perform MLLM contact reasoning to obtain HOI contact information. Finally, aligning the separately reconstructed hands and the articulated object requires up to 5 minutes, yielding the final result. Overall, the full pipeline runtime is dominated by the coarse-to-fine part-wise motion reconstruction, which can be accelerated with a more optimized implementation.

For comparison, RSRD~\cite{kerr2024rsrd} reports similar overall runtime: about 40 minutes to reconstruct and segment the 3D part model from pre-scanned video, roughly 7 minutes for part-motion reconstruction and 4D hand estimation, yet it does not perform any hand-object joint optimization.

\section{Additional Results}

\paragraph{Qualitative Comparison with EasyHOI}

We compare our approach with EasyHOI~\cite{liu2025easyhoi}, a monocular image HOI reconstruction method that also leverages foundation models. Since EasyHOI accepts only single images, we evaluate it frame by frame. For a fair comparison, we use the same foundation models as in our pipeline: WiLoR~\cite{potamias2025wilor} for hand reconstruction and HunYuan3D~\cite{lai2025hunyuan3d} for object shape reconstruction.

Figure~\ref{fig:fig_ours_easyhoi} shows EasyHOI results on ArtHOI-RGBD using single-frame input. EasyHOI generalizes poorly to articulated manipulation because it assumes a fixed object scale and 6-DoF pose, and instead optimizes camera parameters and object pose to fit each image. While this image-based paradigm can be efficient for isolated frames, it clearly fails to produce coherent results on videos.

Moreover, EasyHOI struggles to maintain consistent hand-object alignment across frames. It optimizes contact by considering the entire plausible hand interaction region, which is sufficient for rigid-object grasps, but without specifying contacting fingers, its performance degrades in articulated interactions. The frame-wise reconstruction paradigm also makes video processing computationally infeasible: reconstructing a 100 frame sequence requires roughly 3 hours or more. Finally, EasyHOI assumes a single-hand setting and cannot be easily extended to bimanual scenes without substantial code modifications.

\paragraph{Effect of MLLM Contact Reasoning}

We evaluate the effectiveness of our MLLM-based contact reasoning against a simple rule-based baseline that determines contact via mask intersection. As shown in Table~\ref{tab:mllmvsmask}, while the mask-intersection heuristic shows slightly inferior performance on controlled lab datasets, its accuracy drops substantially on casually captured in-the-wild videos. In contrast, the MLLM leverages broader visual and semantic knowledge, enabling more reliable contact judgments under challenging real-world conditions.

\end{document}